\documentclass{article}

\usepackage{microtype}
\usepackage{graphicx}
\usepackage{subfigure}
\usepackage{booktabs} 

\usepackage{hyperref}
\usepackage[algo2e]{algorithm2e}

\usepackage[accepted]{icml2024}

\usepackage{amsmath}
\usepackage{amssymb}
\usepackage{mathtools}
\usepackage{amsthm}
\usepackage{multirow}
\usepackage{listings}

\usepackage[capitalize,noabbrev]{cleveref}

\theoremstyle{plain}

\theoremstyle{definition}

\theoremstyle{remark}

\usepackage[textsize=tiny]{todonotes}

\icmltitlerunning{Scene Graph Generation Strategy with Co-occurrence Knowledge and Learnable Term Frequency}

\begin{document}

\twocolumn[
\icmltitle{Scene Graph Generation Strategy with Co-occurrence Knowledge and Learnable Term Frequency}



\icmlsetsymbol{equal}{*}

\begin{icmlauthorlist}
\icmlauthor{Hyeongjin Kim}{1}
\icmlauthor{Sangwon Kim}{2}
\icmlauthor{Dasom Ahn}{1}
\icmlauthor{Jong Taek Lee}{3}
\icmlauthor{Byoung Chul Ko}{1}
\end{icmlauthorlist}

\icmlaffiliation{1}{Department of Computer Engineering, Keimyung University, Daegu, South Korea}
\icmlaffiliation{2}{Electronics and Telecommunications Research Institute (ETRI), Daegu, South Korea}
\icmlaffiliation{3}{School of Computer Science and Engineering, Kyungpook National University, Daegu, South Korea}

\icmlcorrespondingauthor{Byoung Chul Ko}{niceko@kmu.ac.kr}

\icmlkeywords{Machine Learning, ICML}

\vskip 0.3in
]



\printAffiliationsAndNotice{}  

\begin{abstract}
Scene graph generation (SGG) is an important task in image understanding because it represents the relationships between objects in an image as a graph structure, making it possible to understand the semantic relationships between objects intuitively. Previous SGG studies used a message-passing neural networks (MPNN) to update features, which can effectively reflect information about surrounding objects. However, these studies have failed to reflect the co-occurrence of objects during SGG generation. In addition, they only addressed the long-tail problem of the training dataset from the perspectives of sampling and learning methods. To address these two problems, we propose CooK, which reflects the Co-occurrence Knowledge between objects, and the learnable term frequency-inverse document frequency (TF-$l$-IDF) to solve the long-tail problem. We applied the proposed model to the SGG benchmark dataset, and the results showed a performance improvement of up to 3.8\% compared with existing state-of-the-art models in SGGen subtask. The proposed method exhibits generalization ability from the results obtained, showing uniform performance improvement for all MPNN models. 
\end{abstract}

\begin{figure}[t]
	\centering
	\includegraphics[width=\columnwidth]{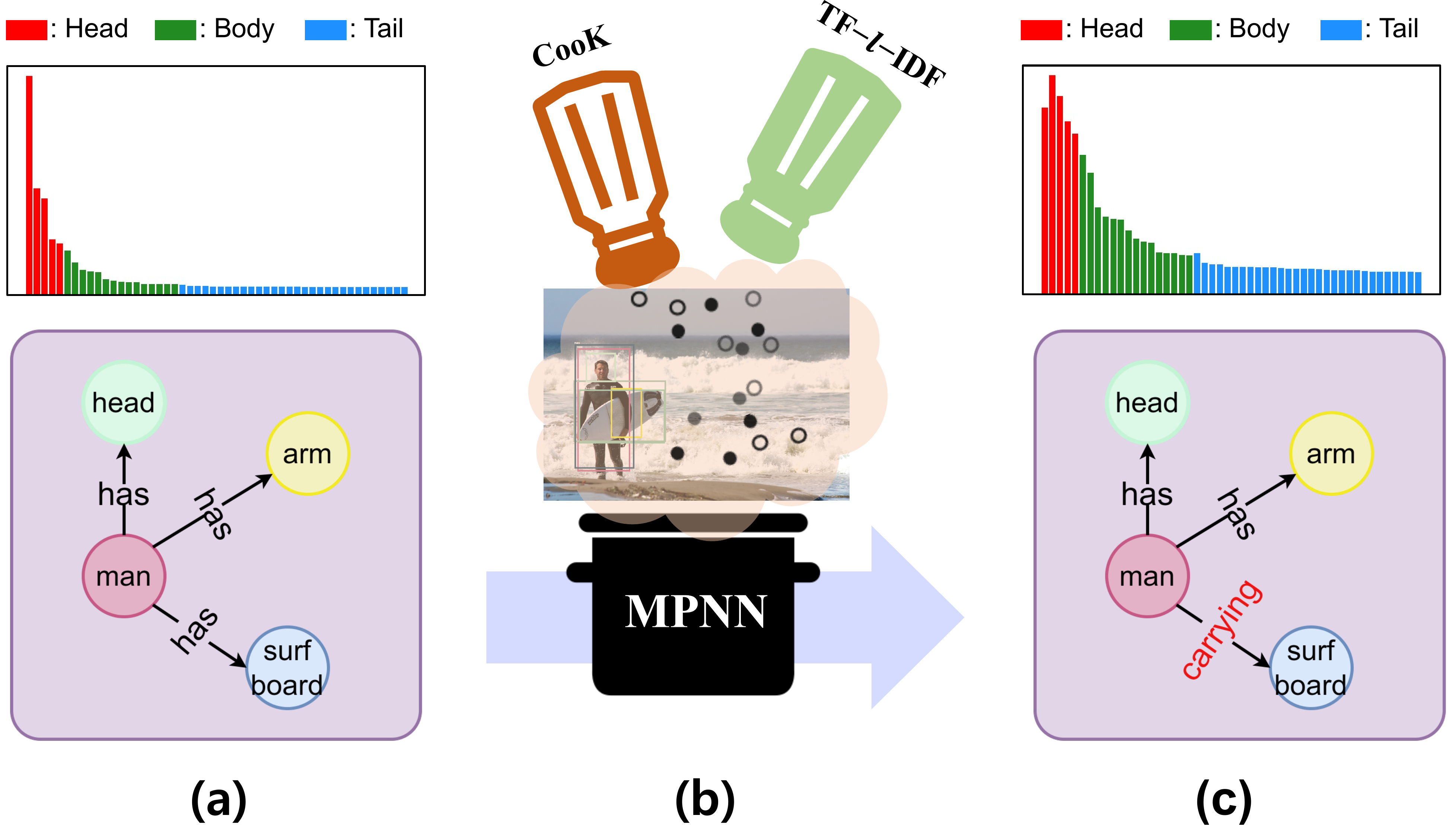}
	\caption{A novel learning recipe for SGG. (a) shows the class distribution and scene graph results of SGG performed using a conventional MPNN-based method. The proposed CooK + TF-$l$-IDF layer can be easily applied to existing MPNN-based models, as shown in (b). By updating the features according to the knowledge of object co-occurrence and the label inverse frequency, as shown in (c), it is possible to generate accurate relations between objects and successfully alleviate the long-tail problem.}
	\label{fig1}
\end{figure}

\section{Introduction}
\label{submission}

Scene graph generation (SGG) is a type of image understanding that infers and interprets the relationships between objects in an image and expresses them as a language graph. SGG has been applied to various computer vision tasks, including image retrieval \cite{iamgeR1, imageR2}, image captioning \cite{imageC1, imageC2}, visual question answering \cite{vqa, vqa1, vqa2, vqa3}, and action recognition \cite{sggActionR1}.
The most common SGG approach is using an object detector \cite{faster, yolo, detr} to infer the relationships between detected objects in an image as a $<$ \verb|subject|, \verb|predicate|, \verb|object|$>$ triplet. This triplet is then represented as a graph with a predicate edge between the subject and object nodes. For example, in the scene graph shown in Figure \ref{fig1}, the subject is ‘man,’ the predicate is ‘carrying,’ and the object is ‘surfboard.’ The relationship between ‘man’ and ‘surfboard’ is represented by the predicate edge ‘carrying,’ which indicates that the ‘a man is carrying a surfboard’.
However, there are many challenges with respect to accurately and effectively inferring the understood content. SGG has evolved to address the challenging problem of inferring relationships between objects. \cite{iamgeR1} first proposed the use of scene graphs for image retrieval and introduced a conditional random field (CRF) model for inferring relationships between objects.
Recently, deep learning methods that utilize graph structures \cite{grcnn, bgnn, hetsgg, edgesgg}, such as graph neural networks (GNNs), have shown excellent performance in SGG. These methods use message-passing neural networks (MPNN) to accurately determine relationships between neighboring objects in a scene, enabling a more effective SGG inference.

However, existing SGG training datasets have a serious long-tail distribution, which can lead to the degradation of fine-grained object relationships and biased predictions toward dominant class labels. For example, a relationship such as ‘walking on’ and ‘carrying’ may be incorrectly predicted as a more dominant class label such as ‘on’ and ‘has.’ Several methods \cite{penet, veto} have been proposed to mitigate this issue.
The methods mentioned above have all contributed to the improvement of performance for successful SGG, but they all have the following limitations: when updating relation features using an MPNN with a graph structure, there is a limitation in that prior knowledge about the mutual correlation between surrounding objects, such as that shown in Figure \ref{fig1} (b) (e.g., person-surfboard, surfboard-wave), is not reflected in this process. State-of-the-art (SoTA) unbiased SGG studies \cite{bgnn, tde, peer, cogtree} are also constrained in their ability to solve this limitation because they focus only on de-biasing between class labels.

To overcome the limitations of existing SGG methods, we propose a novel SGG method that learns object relationships based on Co-occurrence Knowledge (CooK). The proposed SGG learning strategy reflects the co-occurrence information between objects in a scene by calculating the co-occurrence between objects from the training data and applying it to an MPNN. This provides the model with knowledge about the co-occurrence between objects, which has not been carefully considered in previous methods, enabling a more accurate inference of object relationships. In addition, we add a Learnable Term Frequency-Inverse Document Frequency (TF-$l$-IDF) layer to the CooK-based SGG model to alleviate the long-tail problem that exists in scene graph datasets. This layer updates features in a manner that emphasizes the features of the tail class and weakens those of the head classes.
The contributions of the proposed SGG learning strategy are as follows:
\begin{itemize}
	\item \textbf{Reflecting co-occurrence knowledge for accurate relationship inference}: SGG leverages the co-occurrence information between objects in a scene to improve the accuracy of relationship inference. This knowledge, which had been previously neglected by existing methods, helps the model to infer to more precise relationships between objects.
	\item \textbf{Mitigating the long-tail problem through a learnable TF-$l$-IDF layer}: SGG addresses the long-tail problem inherent in scene graph datasets by incorporating a learnable TF-$l$-IDF layer. This layer boosts the features of underrepresented classes (tail classes) while weakening the features of dominant classes (head classes), leading to more balanced and unbiased predictions.
	\item \textbf{Improving SGG performance}: Our research demonstrates that integrating the CooK module into existing SGG models results in significant performance improvements. In addition, experimental results show that it is possible to develop SGG models with better generalization performance by strengthening CooK knowledge with more data.
\end{itemize}

\begin{figure*}[ht]
	\centering
	\includegraphics[width=\textwidth]{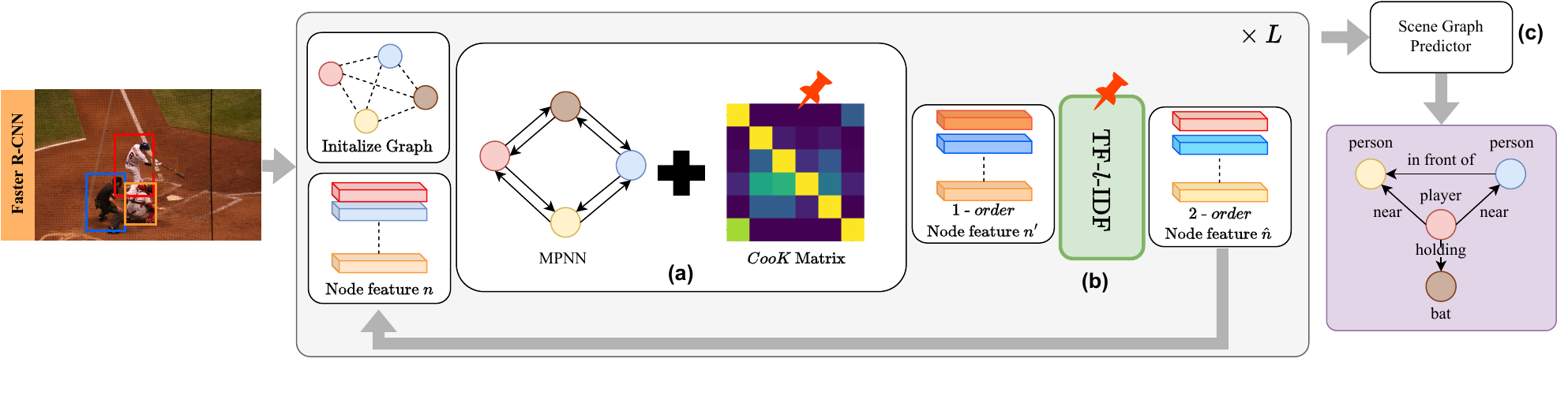}
	\caption{The whole training strategy of our proposed CooK + TF-$l$-IDF method. (a) In the MPNN process, we use the prior knowledge value $CooK(c_j | c_i)$ extracted from the training data to enable learning that reflects CooK. The 1-order node feature $n'$ generated in this way is used as an input to (b) TF-$l$-IDF, which can update features by considering the frequency between labels, to create a 2-order node feature $\hat{n}$. Finally, the 2-order node feature $\hat{n}$ that has undergone $L$ times of (a) and (b) processes is used to generate the final SG through the scene graph predictor in (c).}
	\label{fig2}
\end{figure*}

\section{Related Work}
\subsection{SGG Approaches}
Conventional SGG approaches typically use CNN \cite{lu2016} and RNN \cite{motifs, rnn1} to model object relationships and understand the visual context. These methods typically perform object detection and relation analysis in stages, and they use heuristic rules to generate scene graphs. However, this can lead to the overfitting of complex images or long-tail head classes.

Recently, there has been increased interest in SGG methods that exploit graph structures. Graph-based approaches can effectively reflect surrounding information by updating nodes based on the features of neighboring nodes. A representative graph-based study, Graph-R-CNN \cite{grcnn}, proposed adaptive graph convolution networks that can efficiently update information between objects on top of existing graph convolutional networks.
GPS-Net \cite{gpsnet} proposed direction-aware message passing for node-specific contextual information. BGNN \cite{bgnn} focused on unbiased SGG generation by proposing confidence-aware message propagation. HetSGG \cite{hetsgg} applied unbiased heterogeneous graph structures and updated object-predicate correlated features through the proposed relation-aware message passing, enabling more accurate SGG generation. EdgeSGG \cite{edgesgg} addressed the limitations of existing graph-based SGG models by proposing an edge dual SGG architecture that inverts the roles of each node and edge of the graph. EdgeSGG enables the capture of both object- and edge-centric information, which is essential for generating fine-grained scene graphs. Prior studies on graph-based SGG focused on capturing more accurate relationships between objects and predicates. However, common-knowledge insights have been largely overlooked in SGG, limiting the ability of SGG models to reflect common sense regarding general correlations between objects.

\subsection{Long-Tail Problem Solving}
To improve SGG performance, data-centric approaches that consider the long-tail problem of the training dataset have also attracted attention. The most intuitive approach to addressing the long-tail problem is to design a loss function for long-tail mitigation without modifying the model \cite{knyazev2020graph}. GPS-Net \cite{gpsnet} addressed the long-tail problem in SGG by introducing a reweighted loss. BGNN \cite{bgnn} used a new resampling strategy to construct scene graphs and improve the prediction performance of tail classes. HetSGG \cite{hetsgg} achieved unbiased scene graphs by changing the graph structure from homogeneous to heterogeneous. PE-Net \cite{penet} proposed prototype embedding to make the unbiased feature vectors for each predicate more compact during the SGG process. VETO+MEET \cite{veto} is a mutually exclusive expert learning strategy that is employed for SGG to address long-tail problems. As noted previously, to address the long-tail problem, prior methods predominantly focused on structural modifications to the model. In this study, we introduce a data-centric approach to alleviate long-tail issues.

\subsection{Label Correlation}
Several approaches have been attempted in various fields to incorporate the correlation between labels into learning. For multi-label classification, SSGRL \cite{ssgrl} utilized a graph structure defined by co-occurrence information between labels to improve label classification performance. SALGL \cite{salgl} achieved higher performance multi-label classification than previous studies by simultaneously utilizing co-occurrence knowledge and scene-aware knowledge between labels. In the SGG field, BA-SGG \cite{basgg} enabled more sophisticated scene graph generation by utilizing the co-occurrence information of ‘predicates’ between objects. However, it has the limitation of not considering the co-occurrence knowledge between objects.

\section{Cook + TF-$l$-IDF Recipe}
\subsection{Preliminaries}
\textbf{Scene Graph Generation.} The goal of SGG is to successfully generate a graph $\mathcal{G}=(\mathcal{O},\mathcal{R})$, where $\mathcal{O}$ is a set of objects found in a input image $I$ and $\mathcal{R}$ is a set of relations between them. To generate the graph, we first extract the object set $o_i\in \mathcal{O}$ from the input image $I$ using Faster R-CNN \cite{faster}. Each object $o_i$ is represented as a tuple $(v_i,b_i,c_i)$, where $v_i\in \mathbb{R}^d $ is the visual feature map of $o_i$, $b_i \in [0,1]^4$ is the coordinates of the bounding box of $o_i$, and $c_i\in C $ is the class label of $o_i$. The relation between two objects $o_i$ and $o_j$ is represented as a triplet $r_i = 
\ <o_i,p_{i \rightarrow j},o_j>$, where $p_{i\rightarrow j}$ is the relation feature map that represents the relation from $o_i$ to $o_j$. The feature map can be constructed by concatenating the features of $o_i$ and $o_j$ \cite{squat}, or by updating the visual feature map of the union box of $o_i$ and $o_j$ \cite{bgnn, hetsgg, edgesgg}. A successful SGG aims to learn the relation feature map $p_{i\rightarrow j}$ in a more discriminative manner.

\textbf{Bag of the Word (BoW).} BoW is a method for automatically classifying documents by looking at the distribution of words in a text. BoW considers a term to be relevant if it appears in a document, regardless of the word order or structure of the document. However, BoW does not consider the frequency of words, so term frequency inverse document frequency (TF-IDF) was proposed to address this issue. TF-IDF can reflect the importance of each word by assigning different weights based on the frequency of each word. TF-IDF is calculated as follows:
\begin{equation}
	t_{wd} = TF(n_{td}, n_d) \cdot IDF(N, n_t)
	\label{eq1}
\end{equation}
\begin{equation}
	TF(n_{td}, n_d) = {n_{td} \over n_d},\quad IDF(N, n_t) = \log({N \over n_t})
	\label{eq2}
\end{equation}

where, $n_{td}, n_{d}, n_{t}$ and $N$ means occurrences of word $t$ in document $d$, number of word occurrence in document $d$, number of documents that contain word $t$ and number of documents, respectively.

\subsection{Co-occurrence Knowledge}
Existing SGG learning methods ignore the potential for co-occurrence between objects. Inspired by \cite{cooc}, we propose a CooK that can learn the co-occurrence of objects during the SGG learning stage. CooK is expressed as a matrix, as shown in Figure \ref{fig2} (a); however, for convenience, we refer to it is CooK. For successful CooK-based SGG learning, CooK was extracted from the training dataset. In a training set $D_{train}$  having $K$ images, we count the number of objects with class $i$, $card(oc_i)$ in each image, and count the number of cases where different object class $i$ and $j$ coexist in the same image $card(oc_i \cap oc_j)$. The CooK probability of the object classes $oc_i$ and object class $oc_j$ occurring simultaneously in all images can be calculated as follows:
\begin{equation}
	CooK(c_j|c_i) = {\sum_{k=1}^{K}card_k(oc_i \cap oc_j) \over \sum_{k=1}^{K}card_k(oc_i)}
	\label{eq3}
\end{equation}

\textbf{Advanced CooK.} To obtain more refined knowledge, we adopt two different object recognition datasets, the Visual Genome \cite{vg} and Open Images \cite{oi6}, which are used for SGG learning. This advanced CooK is able to store more extensive knowledge. With this advanced CooK, we can expect significant performance improvements for all SGG subtasks. A detailed discussion of the performance improvement is provided in Section \ref{sec4.7}.

\subsection{Learnable TF-$l$-IDF Layer}
Despite the successful generation of CooK, there remains a long-tail problem. To address this, previous studies have focused primarily on relation classes; however, object classes used for relation inference can also cause serious long-tail problems. Consequently, CooK can be biased in its configuration, which can severely hinder the generalization of overall model learning. To address this issue, we propose a novel method for updating node features by introducing a learnable TF-$l$-IDF layer inspired by TF-IDF scores and adding it to the output of MPNN, as shown in Figure \ref{fig2} (b). The 1-order node features updated by the MPNN and CooK are input to TF-$l$-IDF, which updates them to new 2-order node features. The updated 2-order node features using the TF-$l$-IDF layer reduce the influence of the head class that can occur in the base backbone MPNN and increase the influence of rare body and tail classes.

\textbf{TF-$l$-IDF Layer.} Let the MPNN block be repeated $L$ times, the image batch size be $B$, and the object label set be $\mathcal{O}$. $Z_i^l=\{z_1^l,z_2^l, \cdots z_\mathcal{O}^l \}\in\mathbb{R}^{\mathcal{O}\times d_i}$ is the set of 1-order node feature of the $i$-th object and $d_i$ is the feature dimension. Let the set of $Z_i^l$ of a batch $B$, $X^l=\{Z_1^l,Z_2^l, \cdots Z_B^l \}\in \mathbb{R}^{B\times \mathcal{O}\times d_i}$. $X^l$ is fed to the TF-$l$-IDF layer, and the output is then the updated node feature set $Z_i^{(l+1)}$ :
\begin{equation}
	Z_i^{l+1} = TF\text{-}l\text{-}IDF(X^l)
	\label{eq4}
\end{equation}

The TF-$l$-IDF layer can be expressed as the product of two terms as follows:
\begin{equation}
	TF(X^l|n_{cb}, n_b) \cdot l \text{-}IDF(X^l|B, n_{z_i}; \epsilon, \gamma)
	\label{eq5}
\end{equation}

where $n_{cb}$ is the total number of occurrences of a specific class label $c$ observed in the $i$-th image of the batch, and $n_b$ denotes the total number of occurrences of all object labels in the $b$-th image of the batch. $n_{z_i}$ is the number of images in $z_i$ class. In Equation \ref{eq6}, the TF value represents the frequency of appearance of class $c$ in image $b$.
\begin{equation}
	TF(n_{cb}, n_b) = {n_{cb} \over n_b}
	\label{eq6}
\end{equation}

The $l$-IDF value is the inverse of the TF value.
\begin{equation}
	l\text{-}IDF(B,n_c;\epsilon, \gamma) = \log({B+\epsilon \over n_c + \gamma})
	\label{eq7}
\end{equation}

where $n_{c}$ is the number of images containing class label $c$. To address potential biases introduced during training owing to uneven sampling, we add trainable parameters $\epsilon$ and $\gamma$. These parameters allow for dynamic adjustments to the $\log$ term, thereby minimizing the impact of scenarios in which the body or tail labels are oversampled.
The learnable TF-$l$-IDF layer aims to create a balanced feature representation that addresses the long-tail problem within object classes. By combining the TF with the $l$-IDF, the layer effectively updates the node features, ensuring a more nuanced and unbiased knowledge representation in CooK.

\subsection{Training Strategy}
In this section, we introduce a novel training strategy for SGG that exploits the capabilities of the CooK and TF-$l$-IDF layers. The TF-$l$-IDF layer demonstrates seamless integration with any MPNN-based SGG task, improving model performance in capturing complex relationships within scenes. Figure \ref{fig2} shows the overall pipeline that seamlessly combines CooK and TF-$l$-IDF components. To provide a more concrete illustration of this strategy, we examine the equation using Graph-R-CNN, which is a widely adopted and representative MPNN framework in SGG. The core of MPNN for SGG is determined by the node feature update using the following formula:
\begin{equation}
	z_{o(u\rightarrow v)}^{l+1} = z_{o(u\rightarrow v)}^l + \sigma(z_{o(u)}^l + \sum_{v\in \mathcal{N}(u)}\alpha_{uv}Wz_{o(v)}^l)
	\label{eq8}
\end{equation}

where $z_{o(u)}^l$ denotes the node features of node $u$ included in object $o$ after the $l$-th iteration, with $z_{o(u)}^0$ defined as the initial node feature of node $u$. $\alpha_{uv}$ represents the attention score between nodes $u$ and $v$.
\begin{equation}
	\alpha_{uv} = {\mathrm{exp}(W_{att}u)\over {\mathrm{exp}(W_{att}u)+\mathrm{exp}(W_{att}v)}}
	\label{eq9}
\end{equation}

where $W_{att}$ is the weight used to compute the attention score between $u$ and $v$. To leverage CooK in the MPNN process, Equation \ref{eq8} is replaced with Equation \ref{eq10}.
\begin{multline}
	z_{o(u\rightarrow v)}^{l+1} = z_{o(u\rightarrow v)}^l\ + \\ 
	\sigma(z_{o(u)}^l + \sum_{v\in \mathcal{N}(u)}cook_{u\rightarrow v}\alpha_{uv}Wz_{o(v)}^l)
	\label{eq10}
\end{multline}

Here, $cook_{u \rightarrow v}$ refers to the CooK value that $v$ will occur when object $u$ occurs, and can be easily mapped to the $(u, v)$ values of the CooK calculated in advance. Through this process, CooK is successfully reflected during the MPNN process. For a more detailed explanation of the TF-$l$-IDF-based node feature updating process, refer to Algorithm \ref{Al.1}.

\iftrue
\begin{algorithm}[ht]
	\small{
	\caption{Processing of the TF-$l$-IDF layer}
	\label{Al.1}
	\textbf{Input:}\\
	B: batch size\\
	$\epsilon$ and $\gamma$: parameters for TF-$l$-IDF layer\\
	$n_{cb}$: the total occurrences of a specific class label $c$ observed in the $b$-th image \\
	$n_b$: the total number of occurrences of all object labels in the $b$-th image \\
	$n_{z_i}$: the number of images including $z_i^l$ class \\
	$Z_i^l = \{z_1^l, z_2^l \cdots z_{\mathcal{O}}^l \} \in \mathbb{R}^{\mathcal{O}\times d_i}$ : the set of 1-order node feature of $i$-th image of $l$ times \\
	$X^l = \{Z_1^l, Z_2^l \cdots Z_B^l\} \in \mathcal{R}^{B \times \mathcal{O} \times d_i}$: the set of $Z_i^l$ of a batch B\\
	
	// TF-$l$-IDF score value init.\\
	$n_{cb}$, $n_b$, $n_{z_i} = 0$\\
	
	// TF score set init.\\
	\textbf{TF-score} = $\O$\\
	
	\For {$Z_i^l \in X^l$}{
		$n_b=|Z_i^l|$ \\
		\For {$z_i^l \in Z_i^l$}{
			$n_{cb}=count(b, z_i^l)$ // count $z_i^l$ label in image $b$ \\
			$n_{z_i}$+=$\mathbb{I}_{\{z_i \in b\}}$ // if image $b$ contain label $z_i$\\
			\textbf{TF-score} = \textbf{TF-score} $\cup\ TF(n_{cb}, n_b) \\ \cdot l$-$IDF(B, n_{z_i};\epsilon, \gamma)$
		}
	} 

	\textbf{2-order} $X^l$ = \textbf{TF-score} $\times X^l$ // elemental-wise multiplication} \\ 

	\textbf{Output:} \textbf{2-order} $X^l \in \mathcal{R}^{B \times \mathcal{O} \times d_i}$ 
	
\end{algorithm}
\fi
\subsection{Inference}
Finally, we infer the scene graph $\mathcal{G}=(\mathcal{O},\mathcal{R})$ for the input image $I$ using the successfully trained feature $Z$. The proposed CooK + TF-$l$-IDF learning method applies to all models in the MPNN format. Therefore, we describe the inference process using Graph-R-CNN as an example. The feature $z_{u\rightarrow v}^L$ that has passed through the final $L$ iterations is projected to the final relation class probability vector $p_{u\rightarrow v}$ through a simple linear classifier of weights $W_{rel}$ and softmax function:
\begin{equation}
	p_{u\rightarrow v} = \mathrm{softmax}(W_{rel}z_{u\rightarrow v}^L)
	\label{eq11}
\end{equation}

\textbf{Training Losses.} We use cross-entropy loss to train the MPNN model using the CooK + TF-$l$-IDF layer. The $\mathcal{L}_{obj}$ and $\mathcal{L}_{rel}$ losses for object classification and relation classification are jointly used for the final training as follows:
\begin{multline}
	\mathcal{L}_{obj} = {1 \over |\mathcal{V}|}\sum_{i=1}^{|\mathcal{V}|}\mathcal{L}_{ce}(y_i, \tilde{u}_i),\\ 
	\mathcal{L}_{rel}={1 \over |\mathcal{R}|}\sum_{i=1}^{|\mathcal{R}|}\mathcal{L}_{ce}(s_{u\rightarrow v}, p_{u\rightarrow v}) \\ 
	\mathcal{L} = \mathcal{L}_{obj} + \mathcal{L}_{rel}
	\label{eq12}
\end{multline}

In Equation \ref{eq12}, $y_i$ and $s_{u\rightarrow v}$ represent the ground-truth (GT) for object and relation, respectively. $\tilde{u}$ represents the feature of the $u$ node that has passed through the linear classifier of weights $W_{obj}$ for object classification (e.g., $\tilde{u}_i=W_{obj}u^L_i$). Please refer to the PySGG \cite{pysgg} for detailed training environments of additional models, such as GPS-Net and BGNN.

\begin{table*}[ht]
	\resizebox{\textwidth}{!}{\begin{tabular}{c|c|c|c|c|c|c}
			\toprule
			\multirow{2}{*}{\textbf{Methods}} & \multicolumn{2}{c|}{\textbf{PredCls}} & \multicolumn{2}{c|}{\textbf{SGCls}} & \multicolumn{2}{c}{\textbf{SGGen}} \\ \cline{2-7} 
			& \textbf{mR@ 50 / 100} & \multicolumn{1}{l|}{\textbf{R@ 50 / 100}} & \textbf{mR@ 50 / 100} & \multicolumn{1}{c|}{\textbf{R@ 50 / 100}} & \textbf{mR@ 50 / 100}        & \textbf{R@ 50 / 100}        \\
			\hline \hline
			IMP \cite{imp} & 11.0 / 11.8 & 61.1 / 63.1 & 6.4 / 6.7 & 37.4 / 38.3 & 3.3 / 4.1 & 23.6 / 28.7 \\ 
			KERN \cite{chen2019} &17.7 /19.2 & 65.8 / 67.6 & 9.4 /10.0 & 36.7 / 37.4 & 6.4 / 7.3& 27.1  / 29.8  \\ 
			Motifis \cite{motifs} & 14.6 / 15.8          & 66.0 / 67.9         & 8.0 / 8.5          & 39.1 / 39.9          & 5.5 / 6.8          & 32.1 / 36.9         \\ 
			VCTree \cite{vctree}& 15.4 / 16.6           & 65.5 / 67.4         & 7.4 / 7.9           & 38.9 / 39.8         & 6.6 / 7.7          & 31.8 / 36.1         \\
			G-RCNN \cite{grcnn}& 16.4 / 17.2          & 65.4 / 67.2         & 9.0 / 9.5          & 37.0 / 38.5         & 5.8 / 6.6          & 29.7 / 32.8         \\
			MSDN \cite{mdsn}& 15.9 / 17.5    & 64.6 / 66.6     & 9.3 / 9.7         & 38.4 / 39.8          & 6.1 / 7.2         &  31.9 / 36.6         \\
			Unbiased \cite{unbaised}& 25.4 / 28.7        & 47.2 / 51.6         & 12.2 / 14.0          & 25.4 / 27.9          & 9.3 / 11.1          & 19.4 / 23.2         \\
			GPS-Net \cite{gpsnet}& 15.2 / 16.6          & 65.2 / 67.1         &  8.5 / 9.1         & 37.8 / 39.2         & 6.7 / 8.6          & 31.1 / 35.9         \\ 
			R-CAGCN \cite{rcagcn}& 18.3 / 19.9          & 66.6 / 68.3         & 10.2 / 11.1          & 38.3 / 39.0         & 7.9 / 8.8          & 28.1 / 31.3         \\
			Nice-Motif \cite{nice}& 29.9 / 32.3        & 55.1 / 57.2         & 16.6 / 17.9          & 33.1 / 34.0          & 12.2 / 14.4          & 27.8 / 31.8         \\
			PPDL \cite{ppdl}& 32.2 / 33.3        & 47.2 / 47.6         & 17.5 / 18.2          & 28.4 / 29.3          & 11.4 / 13.5          & 21.2 / 23.9         \\
			RU-Net \cite{runet}& - / 24.2 &  - / 46.9 &  - / 14.6 &  - / 29.0 & - / 10.8 & - / 24.2  \\
			BGNN \cite{bgnn}& 30.4 / 32.9          & 59.2 / 61.3         &  14.3 / 16.5         &  37.4 / 38.5        & 10.7 / 12.6          & 31.0 / 35.8         \\
			IS-GGT \cite{isggt}& 26.4 / 31.9          &   - / -       & 15.8 / 18.9          &  - / -        & 9.1 / 11.3          & - / -         \\
			HetSGG \cite{hetsgg} & 31.6 / 33.5          & 57.8 / 58.9         & 17.2 / 18.7          & 37.6 / 38.7         & 12.2 / 14.4          & 30.0 / 34.6         \\
			HetSGG++ \cite{hetsgg} & 32.3 / 34.5          & 57.1 / 59.4         & 15.8 / 17.7          & 37.6 / 38.5         & 11.5 / 13.5         & 30.2 / 34.5         \\
			PE-Net \cite{penet} & 31.5 / 33.8          & 68.2 / 70.1       &  17.8 / 18.9       & 39.4 / 40.7       & 12.4 / 14.5        &  30.7 / 35.2      \\
			SQUAT  \cite{squat} & 30.9 / 33.4     & - / -       &  17.5 / 18.8      & - / -       & 14.1 / \textbf{16.5}       &  - / -      \\ 
			Transformer+CFA \cite{cfa} & 30.1 / 33.7          & 59.2 / 61.5       & 15.7 / 17.2          & 36.3 / 37.3         & 12.3 / 14.6         & 27.7 / 32.1         \\
			VETO+Rwt \cite{veto} & 33.1 / 35.1& 61.9 / 63.9 & 16.1 / 17.1 & 35.1 / 36.3& 10.0 / 11.7& 26.2 / 30.4  \\ \toprule
			CooK (ours) & 33.7 / 35.8 & 62.1 / 64.2  &  17.5/ 18.6 &  39.1 / 40.0  & 12.6 / 14.9 & 30.1 / 34.6   \\
			TF-$l$-IDF (ours) & 33.6 / 35.8  & 61.7 / 63.4   &  18.5 / 19.4 &  38.4 / 39.8 & 12.8 / 15.0  & 29.3 / 32.6  \\
			CooK + TF-$l$-IDF (ours) & \textbf{35.4 / 37.2} & 60.4 / 62.3 &  \textbf{19.1 / 20.3} &  36.4 / 37.6 & \textbf{14.2} / 16.3 & 27.7 / 32.7 \\ \bottomrule         
	\end{tabular}}
	\caption{Performance comparison with the SoTA SGG methods on the VG dataset.}
	\label{table1}
\end{table*}

\section{Experiment}
\subsection{Datasets}
To verify the performance of the proposed CooK + TF-$l$-IDF method, experiments were conducted on the following two datasets: Visual Genome \cite{vg} and Open Images \cite{oi6}.

\textbf{Visual Genome (VG).} The VG dataset consists of 108k images, with 150 object class labels and 50 relation labels. The dataset was divided into 70\% training data and 30\% test data, and preprocessing was performed according to a previous study \cite{imp}. 

\textbf{Open Images (OI).} The OI dataset consists of 133k images, with 301 object class labels and 31 relation labels. A total of 126,368 images were used for training; 1,813 and 5,322 images were used for validation and testing, respectively.

\subsection{Evaluation Metrics}
\textbf{Visual Genome (VG).} To evaluate the performance of SGG on the VG dataset, we report the performance of three subtasks: Predicate Classification (PredCls), Scene Graph Classification (SGCls), and Scene Graph Generation (SGGen), which have been used in previous studies \cite{lyu2022fine, tde, vctree}. PredCls are given labels and bounding box information for the object. SGCls is given only the bounding box information for the object, and SGGen uses only the values detected by the object detector without the GT label for the object. Following previous studies, we used recall@K (R@K) and mean recall@K (mR@K) as the primary evaluation metrics.

\textbf{Open Images (OI).} Following previous studies, we report the performance of the Recall@50 (R@50), weighted mean AP of relation $\mathrm{\mathbf{wmAP}}_{rel}$, and weighted mean AP of phrase $\mathrm{\mathbf{wmAP}}_{phr}$ metrics \cite{bgnn}, which are the main metrics used to measure performance on the OI dataset. In addition, we reported the final score $\mathrm{\mathbf{score}}_{wtd}$, which was calculated as the weighted sum of the following three indicators:
\begin{multline}
	\mathrm{\mathbf{score}}_{wtd} = 0.2 \times \mathrm{R@}50 + 0.4 \times \mathrm{\mathbf{wmAP}}_{rel} + \\ 
	0.4 \times \mathrm{\mathbf{wmAP}}_{phr}
	\label{eq13}
\end{multline}

\subsection{Implementation Details}
All of the experiments were conducted on a private machine equipped with two Intel(R) Xeon(R) CPUs, that is, a Gold 6230R CPU @ 2.10 GHz; 128 GB RAM, and an NVIDIA RTX 3090 GPU. To detect objects in the image, we adopted the Faster R-CNN \cite{faster} with ResNeXt-101-FPN \cite{resxt}. GloVe \cite{glove} was used to word embedding method. 

\begin{table*}[t]
	\begin{center}
		\begin{tabular}{c|c|c|c|c|c}
			\toprule
			\textbf{Methods}	&\textbf{mR@50} & \textbf{R@50} &  $\mathrm{\mathbf{wmAP}}_{rel}$ & $\mathrm{\mathbf{wmAP}}_{phr}$ & $\mathrm{\mathbf{score}}_{wtd}$ \\ \hline \hline
			RelDN \cite{reldn} & 37.2 & 75.3 & 32.2 & 33.4 & 42.0 \\
			VCTree \cite{vctree} & 33.9 & 74.1 & 34.2 & 33.1 & 40.2 \\
			G-RCNN \cite{grcnn} & 34.0 & 74.5 & 33.2 & 34.2 & 41.8 \\
			Motifs \cite{motifs} & 32.7 & 71.6 & 29.9 & 31.6 & 38.9 \\
			Unbiased \cite{unbaised} & 35.5 & 69.3 & 30.7 & 32.8 & 39.3 \\
			GPS-Net \cite{gpsnet}& 38.9 & 74.7 & 32.8 & 33.9 & 41.6 \\
			BGNN \cite{bgnn}& 40.5 & 75.0 & 33.5 & 34.1 & 42.1 \\
			RU-Net \cite{runet}& - & 76.9 & 35.4 & 34.9 & 43.5 \\
			HetSGG \cite{hetsgg}& 42.7 & 76.8 & 34.6& 35.5 & 43.3 \\
			HetSGG++ \cite{hetsgg}& 43.2 & 74.8 & 33.5 & 34.5 & 42.2 \\
			PE-Net \cite{penet}& - & 76.5 & \textbf{36.6} & 37.4 & 44.9 \\ \hline
			CooK (ours)& 42.9 & 75.5 & 34.6 & 36.4 & 43.5 \\ 
			TF-$l$-IDF (ours)& 43.3 & 76.5 & 35.4 & 36.8 & 44.2 \\ 
			CooK + TF-$l$-IDF (ours)& \textbf{43.8} & \textbf{77.0} & \textbf{36.6} & \textbf{37.6} & \textbf{45.1} \\ \bottomrule
		\end{tabular}
	\end{center}
	\caption{Performance comparison with the SoTA methods on OI dataset.}	
	\label{table2}
\end{table*}

\begin{table}[ht]
	\begin{center}
		\resizebox{\columnwidth}{!}{%
			\begin{tabular}{c|cccc}
				\toprule
				\multicolumn{1}{c|}{\multirow{2}{*}{\textbf{Method}}} & \multicolumn{4}{c}{\textbf{SGGen}} \\ 
				\multicolumn{1}{c|}{}                   & \textbf{mR@50}     & \textbf{mR@100}     & \textbf{R@50}  & \textbf{ R@100}   \\ \hline \hline
				G-RCNN \cite{grcnn}&   5.8   &  6.6    &  29.7     & 32.8\\
				G-RCNN + ours&    7.1   &  8.7     &  30.1     & 33.2\\ \hline
				GPS-Net \cite{gpsnet}&   6.7    &  8.6    &   31.1   & 35.9\\
				GPS-Net + ours&   8.3    &  10.6     &   33.5    & 37.4\\\hline
				BGNN \cite{bgnn}&    10.7   &  12.6     &   31  & 35.8\\ 
				BGNN+ ours&    11.4   &  14.2     &   29.8  & 34.6\\ \hline
				\textbf{Mean improv.(\%)}&    {\textbf{17.6}}$\uparrow$   &   {\textbf{22.6}}$\uparrow$    &   {\textbf{1.6}}$\uparrow$  & {\textbf{0.7}}$\uparrow$\\ \bottomrule
		\end{tabular}}
	\end{center}
	\caption{Performance changes when the proposed CooK+TF-$l$-IDF are applied to various MPNN based models in the VG dataset.}
	\label{table3}
\end{table}

\subsection{Increasing Performance Using CooK + TF-$l$-IDF}
\textbf{Visual Genome.} Table \ref{table1} compares the performances of the proposed CooK + TF-$l$-IDF and SoTA models. As shown in Table \ref{table1}, our proposed method significantly improves the performance of most evaluation metrics. For CooK, which includes the co-occurrence knowledge between objects, PredCls showed a performance improvement of 2.1\% / 2.3\% on mR@50 / 100 because the object GT was reflected in CooK. In addition, SGCls and SGGen each improved the performance by 0.3\% / 0.1\% and 0.4\% / 0.5\% on mR@50 / 100, respectively, but the performance improvements were smaller than those of PredCls. This is because neither method reflects CooK information and uses object labels directly predicted by the model. When only the TF-$l$-IDF layer was used, similar levels of performance improvement were observed for all three subtasks. In particular, as evidence of the performance improvement for the tail classes, which is the role of TF-$l$-IDF, the performance improvement of R@K was larger than that of mR@K. Finally, the largest performance improvement was observed when using information for both CooK and TF-$l$-IDF. This is the result of considering the knowledge of both object co-occurrence and class balance, which shows that more accurate SGG is possible through these considerations.

\begin{table}[ht]
	\begin{center}
		\begin{tabular}{c|ccc}
			\toprule
			\multicolumn{1}{c|}{\multirow{2}{*}{Learnable}} & \multicolumn{3}{c}{\textbf{PredCls}} \\ 
			\multicolumn{1}{c|}{}                   & \textbf{mR@20}     & \textbf{mR@50}     & \textbf{mR@100}  \\ \hline \hline
			$w/o$&  26.8    &  31.6    &  33.4   \\
			$w$&   \textbf{29.9}    &  \textbf{33.6}    &  \textbf{35.8}    \\ \bottomrule
		\end{tabular}
		
		\begin{tabular}{c|ccc}
			\toprule
			\multicolumn{1}{c|}{\multirow{2}{*}{Learnable}} & \multicolumn{3}{c}{\textbf{SGCls}} \\ 
			\multicolumn{1}{c|}{}                   & \textbf{mR@20}     & \textbf{mR@50}     & \textbf{mR@100}  \\ \hline \hline
			$w/o$&  12.2    & 15.1     &  16.3   \\
			$w$& \textbf{16.1}      &  \textbf{18.5 }   &  \textbf{19.4}    \\ \bottomrule
		\end{tabular}
		
		\begin{tabular}{c|ccc}
			\toprule
			\multicolumn{1}{c|}{\multirow{2}{*}{Learnable}} & \multicolumn{3}{c}{\textbf{SGGen}} \\ 
			\multicolumn{1}{c|}{}                   & \textbf{mR@20}     & \textbf{mR@50}     & \textbf{mR@100}  \\ \hline \hline
			$w/o$&  8.3    &  11.1    &  13.2   \\
			$w$&  \textbf{ 9.7}    &   \textbf{12.8}   & \textbf{15.0}   \\ \bottomrule
		\end{tabular}
	\end{center}
	\caption{Performance changes depending on the with ($w$) or without ($w/o$) learnable parameters in the TF-$l$-IDF layer.}
	\label{table4}
\end{table}

\textbf{Open Images.} To verify the generalized performance improvement of the proposed method, we conducted experiments on the SGGen task using the OI dataset, as listed in Table \ref{table2}. Table \ref{table2} presents the results of the performance evaluation of the OI dataset. As with the VG dataset, our CooK+TF-$l$-IDF method showed a performance improvement of 0.2\% over the SoTA models. In particular, unlike PE-NET, which requires training with prototypes, our proposed method can achieve a higher performance improvement using CooK and TF-$l$-IDF extracted from the training data.

\begin{figure*}[ht]
	\centering
	\includegraphics[width=\textwidth]{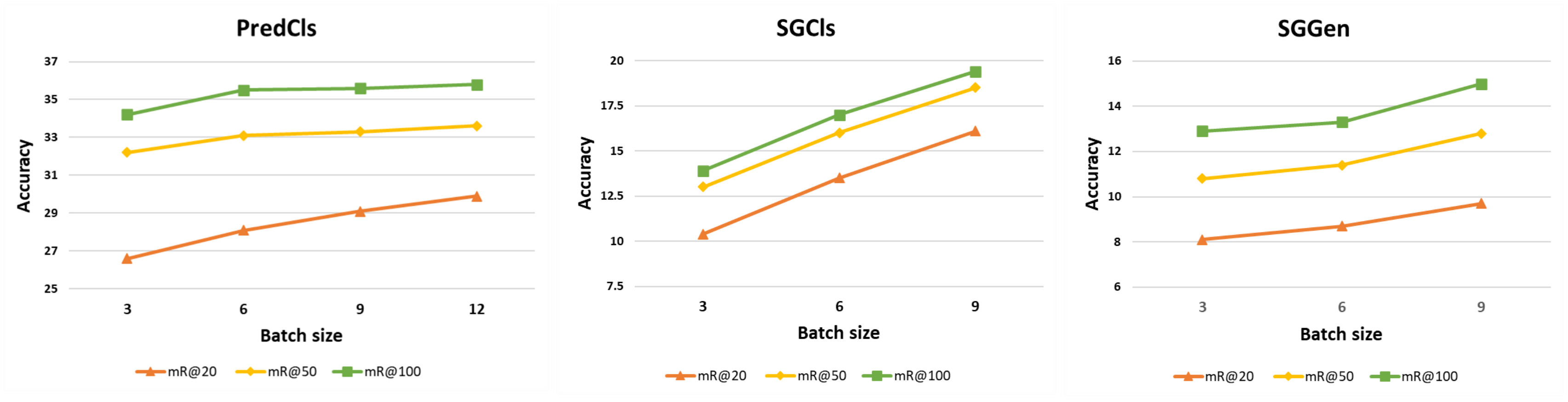}
	\caption{Difference in TF-$l$-IDF performance according to the batch size. As the proposed TF-$l$-IDF is performed in batches, it can be confirmed that the performance increases proportionally as the batch size increases.}
	\label{fig3}
\end{figure*}

\textbf{CooK + TF-$l$-IDF with MPNN-based models.} To verify the generalized performance improvement of our proposed method, we examined the changes in the SGGen performance when CooK and TF-$l$-IDF layers were applied to representative MPNN-based SGG methods \cite{grcnn, gpsnet, bgnn}. As shown in Table \ref{table3}, we can confirm that there was a performance improvement in all MPNN-based models. This shows that the proposed method can be broadly applied to MPNN-based SGG tasks and can achieve high generalization performance.

\begin{table}[ht]
	\begin{center}
		\begin{tabular}{c|ccc}
			\toprule
			\multicolumn{1}{c|}{\multirow{2}{*}{CooK-Type}} & \multicolumn{3}{c}{\textbf{SGGen}} \\ 
			\multicolumn{1}{c|}{}                   & \textbf{mR@20}     & \textbf{mR@50}     & \textbf{mR@100}  \\ \hline \hline
			One CooK for VG& 11.1 &  14.2    &  16.3   \\
			Advanced&  \textbf{11.4}   & \textbf{14.9}  & \textbf{17.1}  \\ \bottomrule
		\end{tabular}
	\end{center}
	\caption{Performance changes when using Advanced CooK, a prior knowledge collected from a wider variety of environments. Advanced CooK achieved a higher performance than when it was not used. Because the OI dataset has more object labels than the VG dataset, $Map:OI\rightarrow VG$  was excluded as many cases occurred where mapping was not possible. }
	\label{table5}
\end{table}

\subsection{Ablation Studies}
\textbf{Use of Learnable TF-$l$-IDF.} The most significant contribution of the proposed TF-$l$-IDF is the learnable design of the IDF(,) function, which calculates the inverse document frequency. To compare the extent of the performance improvement, we compared the performance of the structure with the learnable design of TF-$l$-IDF and the performance of the structure without it. As can be seen in Table \ref{table4}, the use of learnable parameters led to an average 2.4\% performance improvement when compared to the case where they were not used. This is because learnable parameters can mitigate the cases in which a specific label is oversampled during training.

\textbf{Difference in TF-$l$-IDF Performance According to Batch Size.} As discussed in the previous section, TF-$l$-IDF is highly dependent on the batch size. Figure \ref{fig3} depicts the performance of the TF-$l$-IDF layer on the three subtasks of PredCls, SGCls, and SGGen on the VG dataset according to the batch size. As shown in the figure, the performance gradually increased for all subtasks as the batch size increased. Therefore, it is necessary to increase the batch size to perform more sophisticated feature updates.

\subsection{Long-tail Alleviation}
\begin{figure}[th]
	\resizebox{\columnwidth}{!}{%
	\includegraphics[width=\textwidth]{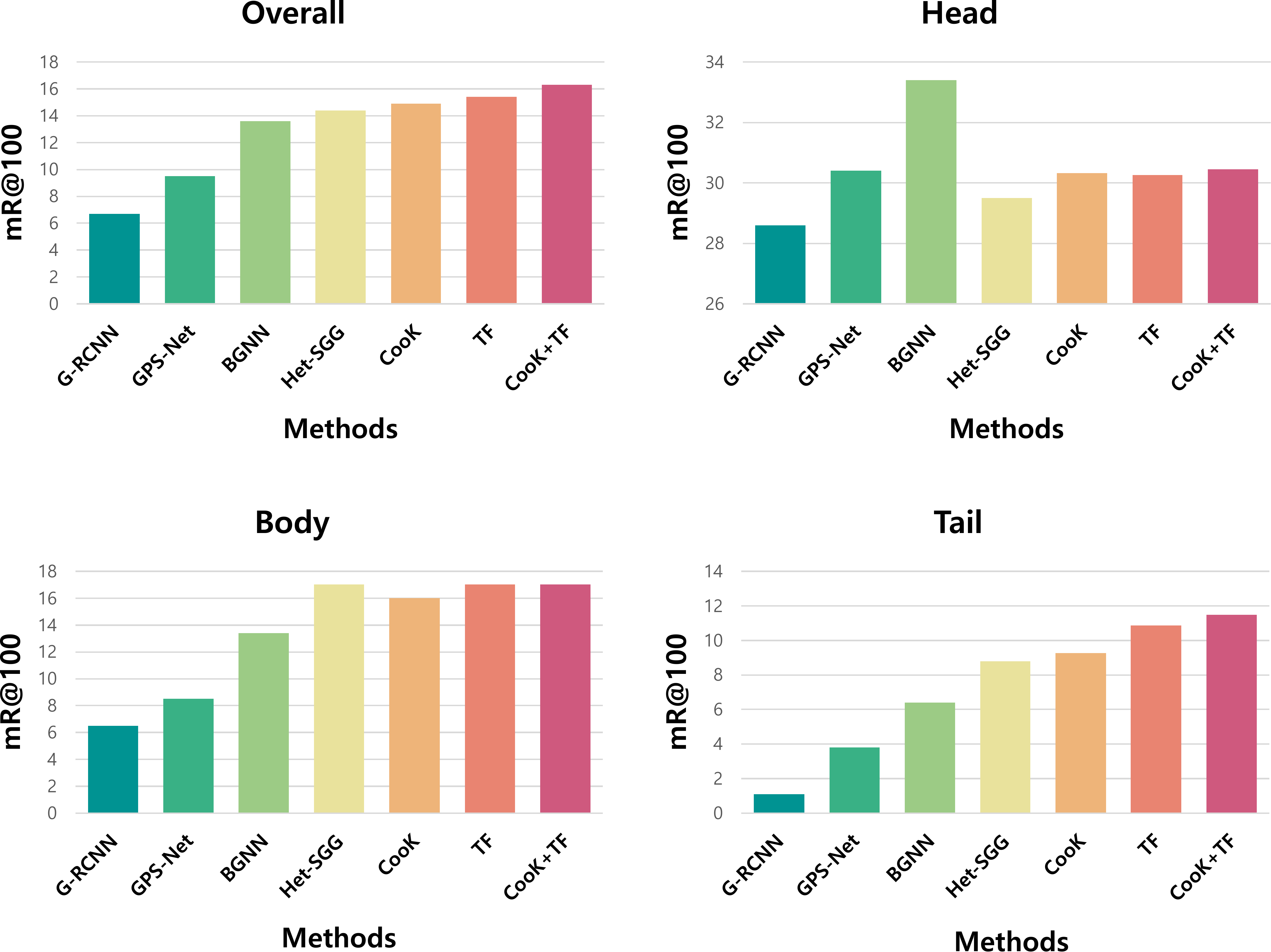}} 
	\caption {TF-$l$-IDF effect on long-tail problem. The proposed TF-$l$-IDF successfully reduces the mR@100 for common labels in head and focuses more on rare labels in body and tail. }
	\label{fig4}
\end{figure}

In this experiment, we analyzed the effects of the proposed learning method on long-tail problem mitigation. Figure \ref{fig4} shows the change in mR@100 for each class when the proposed method was applied. As illustrated in the figure, the mR@100 value for the head decreased, whereas those for the body and tail parts increased significantly. This demonstrates that CooK’s ‘knowledge of object co-occurrence’ and TF-$l$-IDF's ‘feature update’ were successfully applied to each class part.

\subsection{Advanced CooK}
\label{sec4.7}
Human knowledge has become more generalized and reliable owing to extensive experience and activities. To verify the applicability of this human knowledge paradigm to CooK, we generated an advanced CooK based on additional datasets and applied it to the model. In this experiment, we combined two CooKs from the VG and OI datasets to create an advanced CooK. The mapping function $Map:OI\rightarrow VG$ for the combination was hand-crafted. Table \ref{table5} shows the performance results when using advanced CooK. Similar to the general improvement effect of knowledge, the advanced CooK achieved a higher performance than individual CooK. This demonstrates that CooK can improve the performance if the task uses knowledge obtained from similar datasets.

\section{Limitation and Future work}
The proposed paper successfully improves the performance of MPNN-based SGG models. However, exploring the application of CooK and TF-IDF to non-MPNN models could present a novel approach for SGG. As future work, we intend to investigate the feasibility and effectiveness of integrating the proposed CooK and TF-IDF methods into other approaches beyond MPNN-based models.

\section*{Conclusion}
In this study, we proposed a CooK and TF-$l$-IDF layer that can solve co-occurrence knowledge and long-tail problems. Our proposed method has a significant advantage in that it can improve the performance of SGG tasks because it can be easily applied without significantly changing the existing model. In addition, by performing experiments using advanced CooK, we verify that this study realized a new approach for generating more general knowledge matrices. However, limitation is that it can only be applied to existing MPNN-based models. In addition, CooK learning is currently applicable only to supervised learning; therefore, it is difficult to apply it to foundation models conducted with self-supervision. In future research, we plan to study SGG, which enables CooK to learn the relationships between objects on its own and is applicable regardless of the SGG model.

\section*{Acknowledgments}
This work was supported by Korea Institute for Advancement of Technology (KIAT) grant funded by the Korea Government (MOTIE) (P0020536, HRD Program for Industrial Innovation) and Basic Science Research Program through the National Research Foundation of Korea (NRF) funded by the Ministry of Education (2022R1I1A3058128).
 
\section*{Impact Statements}
This research presents work whose goal is to advance the field of Machine Learning. There are many potential societal consequences of our work, none which we feel must be specifically highlighted here.


\bibliography{CooK}
\bibliographystyle{icml2024}

\newpage
\appendix
\onecolumn
\begin{center}
	{\Large Scene Graph Generation Strategy with Co-occurrence Knowledge and \\ Learnable Term Frequency} \\
	{\large \textbf{Appendix}}
\end{center}

\section{Training Configuration}

\begin{table}[ht]
	\setlength{\tabcolsep}{10pt}
	\renewcommand{\arraystretch}{1.5}
	\begin{center}
			\begin{tabular}{c|cc|c}
			\toprule
			\multirow{3}{*}{Hyperparameters} & \multicolumn{3}{c}{Datasets}         \\ \cline{2-4} 
			& \multicolumn{2}{c|}{VG} & OI    \\ \cline{2-4}
			& PredCls         & \multicolumn{1}{c|}{SGCls, SGGen}         & SGGen     \\ \hline \hline
			LR&    0.008       &    0.008       &  0.008     \\
			LR decay&    WarmupMultiStepLR &      WarmupMultiStepLR &   WarmupMultiStepLR \\
		Weight decay&     $5 \times 10^{-5}$        &     $5 \times 10^{-5}$   &   $5 \times 10^{-5}$\\
		Iteration&    49,500       &      49,500   & 40,000 \\
		Batch size&    12       &     9   &   9\\
		\# layers&      4     &      4   &  4\\
		Object dim&     128      &   128   &   128  \\
		Relation dim&    128       &     128  &  128    \\ \bottomrule
		\end{tabular}
	\end{center}
	\caption{Model configurations for the benchmark datasets used in the experiments.}
	\label{tab.a1}
\end{table}

In order to ensure the reproducibility of the proposed model and the reliability of the experiments, we provide detailed hyperparameter values employed in the experiment. Table \ref{tab.a1} shows the model configurations on each benchmark dataset.

\newpage
\section{CooK Visualization}
	\begin{center}
		\begin{figure}[ht]
		
			\includegraphics[width=\textwidth]{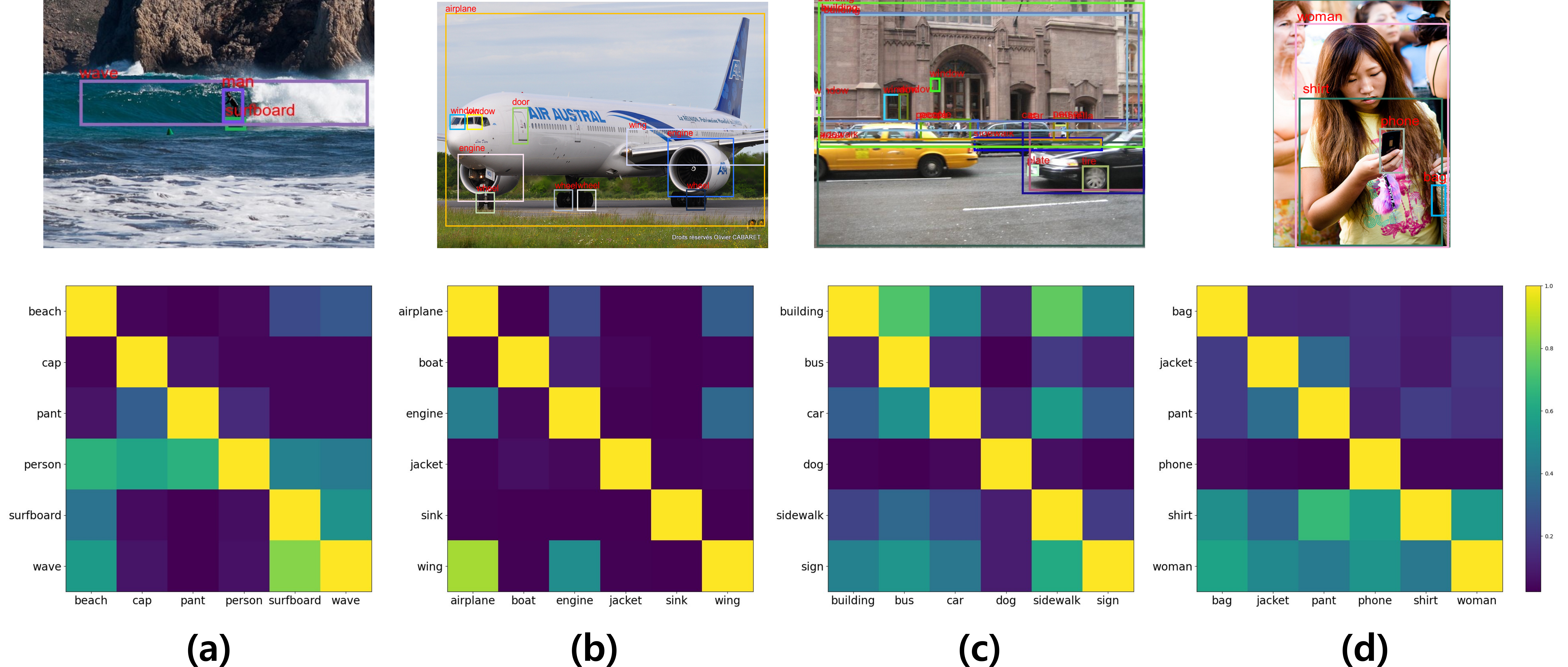}
			\caption{Visualization of CooK matrix. CooK above is a visualization of CooK that reflects object co-occurrence in VG dataset.}		\label{fig.a1}
		\end{figure}
	\end{center}

We visualize CooK to make it easier to explain how well the proposed CooK represents the correlation between objects. As you can see in the figure, it clearly reflects the relationship between objects that are closely related in the real-world of CooK. When looking at the relationship between the `surfboard' and `wave' in the case of (a), if there is a `surfboard', it can be seen that it occurs together with high probability of `wave' and `beach'. Also, `beach' and `surfboard' show lower probability of co-occurrence than `surfboard' and `beach'. This is, of course, an accurate reflection of the fact that the `surfboard' does not necessarily exist in the `beach'. In other (b), (c), and (d), it can be seen that CooK reflects real-world co-occurrence knowledge well.

\newpage
\section{Long-Tail Mitigation}

\begin{figure}[ht]
	\centering
	\includegraphics[width=\textwidth]{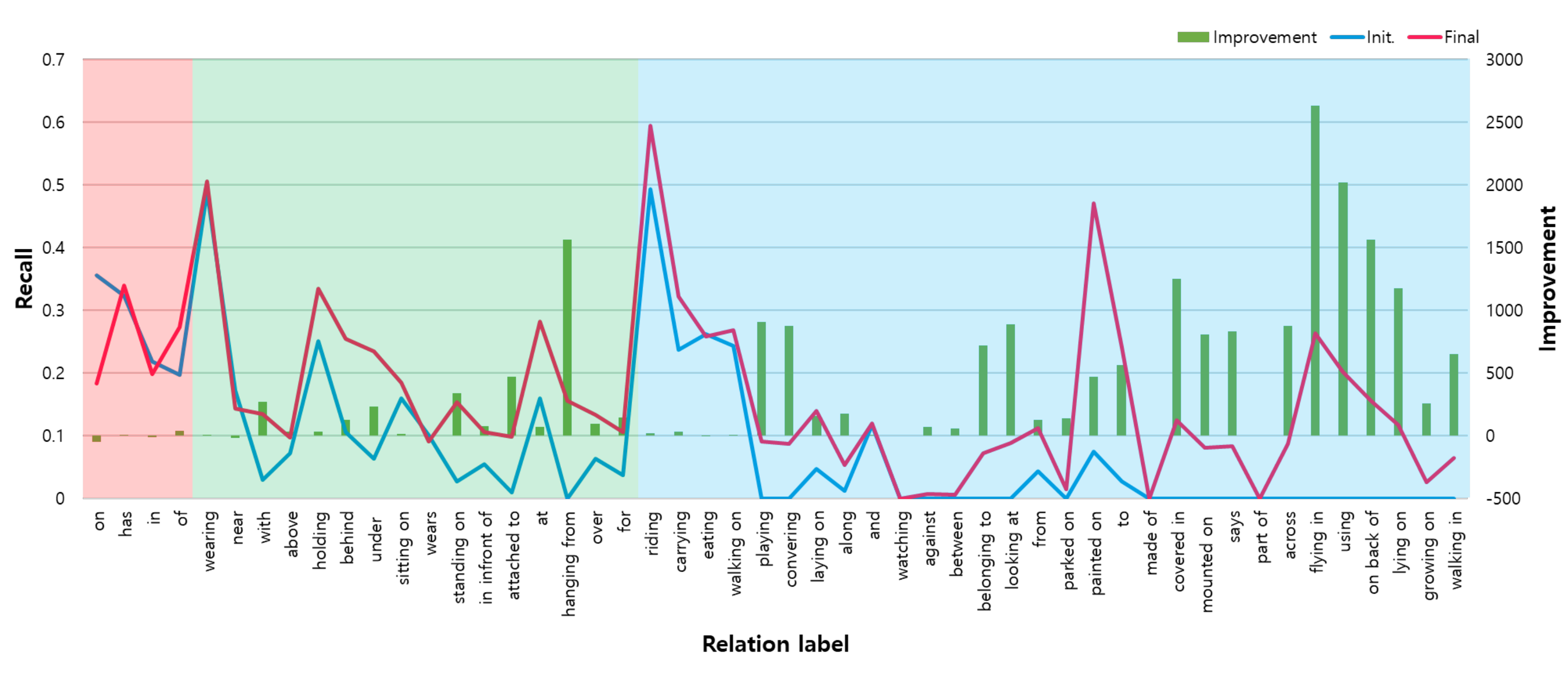}
	\caption{Mitigation of long-tail problems after training.}
	\label{fig.a2}
\end{figure}

To add clarity to the long-tail problem resolution of the proposed method, we additionally report the recall (R@100) value and degree of improvement of the relocation labels in the SGGen task as learning progresses. The red area in the figure represents head classes, the green area represents body classes, and the blue area represents tail classes. The line graph represents the recall change of the relocation label according to model learning (right axis). As you can see in the figure, Figure \ref{fig.a2} the change in tail is more dynamic than the change in head or body. This is more evident when you check the bar graph indicating the degree of improvement (left axis). It can be seen that the proposed TF-$l$-IDF decreases the frequency of the common class and increases the frequency of the rare class dramatically.

\newpage
\section{Scene Graph Visualization}

\begin{figure}[ht]
	\centering
	\includegraphics[width=\textwidth]{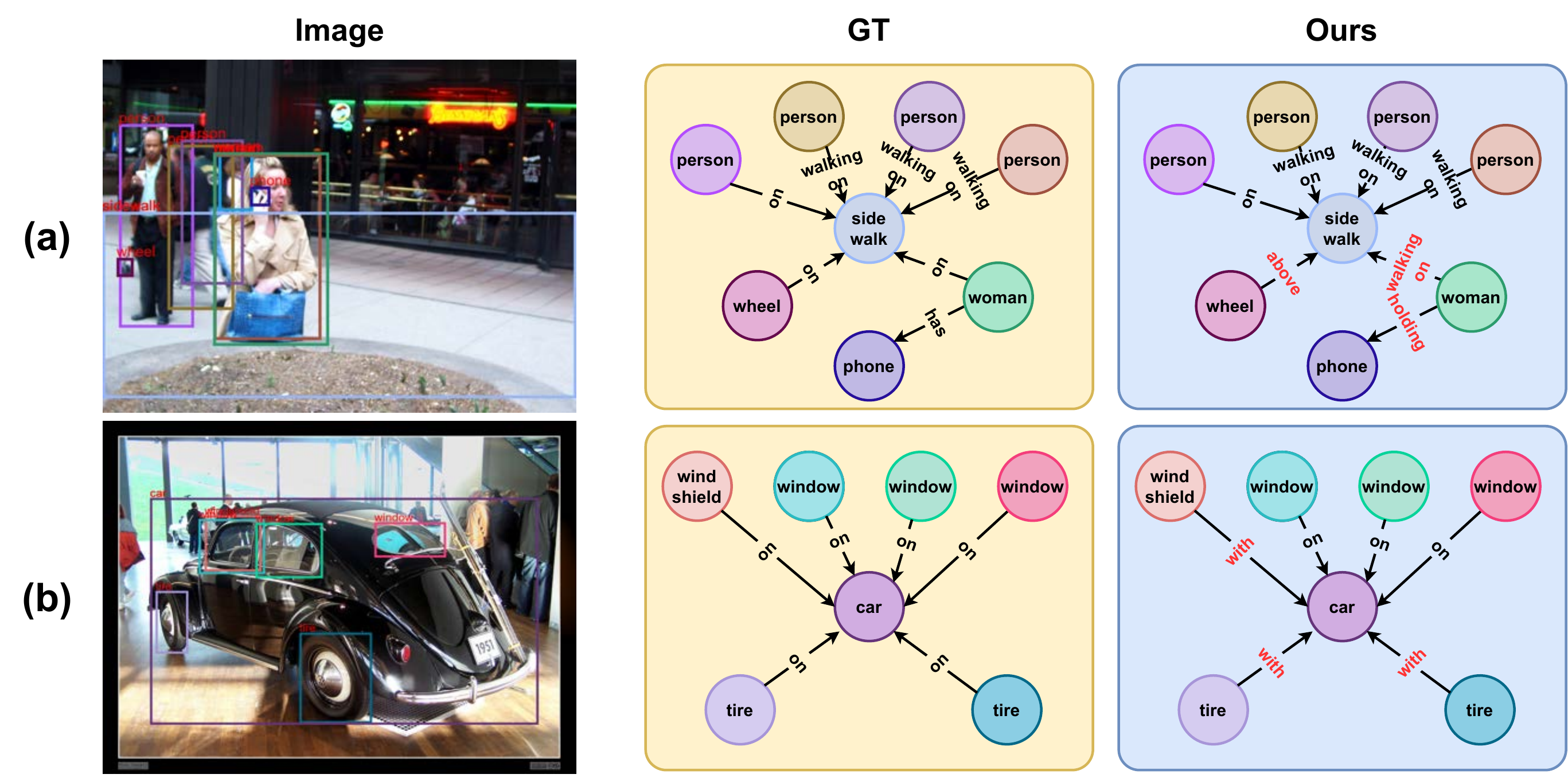}
	\caption{Visualization results of scene graph in PredCls subtask on VG dataset.}
	\label{fig.a3}
\end{figure}

To prove the qualitative results of the proposed CooK + TF-$l$-IDF, we introduce the visualization results in PredCls subtask. As you can see in the Figure \ref{fig.a3}, the proposed method makes more accurate relation predictions. As shown in Figure \ref{fig.a3} (a) and (b), we can see that our method avoids prediction to the dominant head class and performs prediction to a more delicate class.

\newpage
\section{TF-$l$-IDF Code}

\definecolor{codegreen}{rgb}{0,0.6,0}
\definecolor{codegray}{rgb}{0.5,0.5,0.5}
\definecolor{codepurple}{rgb}{0.58,0,0.82}
\definecolor{backcolour}{rgb}{0.95,0.95,0.92}

\lstdefinestyle{mystyle}{
	backgroundcolor=\color{backcolour},   
	commentstyle=\color{codegreen},
	keywordstyle=\color{magenta},
	numberstyle=\tiny\color{codegray},
	stringstyle=\color{codepurple},
	basicstyle=\ttfamily\footnotesize,
	breakatwhitespace=false,         
	breaklines=true,                 
	captionpos=b,                    
	keepspaces=true,                 
	numbers=left,                    
	numbersep=5pt,                  
	showspaces=false,                
	showstringspaces=false,
	showtabs=false,                  
	tabsize=2
}

\lstset{style=mystyle}

\iftrue
\begin{lstlisting}[language=Python]
import torch
import torch.nn as nn
import math
import numpy as np

class TfIdfLayer(nn.Module):
	def __init__(self, epsilon=0.0, gamma=0.0, bias=False):
		super(TfIdfLayer, self).__init__()
		self.epsilon = epsilon
		self.gamma = gamma
		self.bias = bias

		if self.bias:
			self.epsilon = nn.Parameter(torch.randn(1), requires_grad=True)
			self.gamma = nn.Parameter(torch.randn(1), requires_grad=True)

	def forward(self, x, labels):

		num_img = len(labels)
		t_id_list = []
		
		for label in labels:
			for ob_label in label.extra_fields['pred_labels']:
				tf_idf = self.tfidf(ob_label, label.extra_fields['pred_labels'], num_img, labels)
				t_id_list.append(tf_idf)
		
		weighted_x = torch.tensor(t_id_list).unsqueeze(-1).cuda() * x
	
		return weighted_x

	def tf(self, t, d):
		return torch.count_nonzero(torch.where(d == t, True, False)).item()

	def idf(self, t, N, docs):
		ni = 0
		for doc in docs:
			ni += t in doc.extra_fields['pred_labels']
	
		if self.bias:
			return math.log((N + self.epsilon) / (ni + 1 + self.gamma))
		else:
			return math.log(N / (ni + 1))
	
	def tfidf(self, t, d, N, docs):
		return self.tf(t, d) * self.idf(t, N, docs)
\end{lstlisting}
\fi


\end{document}